\documentclass{llncs}

\usepackage{graphicx}
\usepackage{caption}
\usepackage{subcaption}
\usepackage{amsmath}
\usepackage{hyperref}
\usepackage{placeins}

\begin{document}

\title{Evolving Robots on Easy Mode: Towards a Variable Complexity Controller for Quadrupeds\thanks{This work is partially supported by The Research Council of Norway under grant agreement 240862.}}

\author{T{\o}nnes F. Nygaard\inst{1} \and
        Charles P. Martin\inst{1} \and
        Jim Torresen\inst{1} \and
        Kyrre Glette\inst{1}
}


\institute{University of Oslo, Norway\\ \email{tonnesfn@ifi.uio.no}}

\maketitle

\begin{abstract}
    The complexity of a legged robot's environment or task can inform how specialised its gait must be to ensure success. Evolving specialised robotic gaits demands many evaluations---acceptable for computer simulations, but not for physical robots. For some tasks, a more general gait, with lower optimization costs, could be satisfactory. In this paper, we introduce a new type of gait controller where complexity can be set by a single parameter, using a dynamic genotype-phenotype mapping. Low controller complexity leads to conservative gaits, while higher complexity allows more sophistication and high performance for demanding tasks, at the cost of optimization effort. We investigate the new controller on a virtual robot in simulations and do preliminary testing on a real-world robot. We show that having variable complexity allows us to adapt to different optimization budgets. With a high evaluation budget in simulation, a complex controller performs best. Moreover, real-world evolution with a limited evaluation budget indicates that a lower gait complexity is preferable for a relatively simple environment.
    \keywords{Evolutionary Robotics \and Real-world Evolution \and Legged Robots}
\end{abstract}

%
%

\section{Introduction}
Robots are used in more and more demanding and changing environments.
Being able to adapt to new situations, unexpected events, or even damage to the robot itself can be crucial in many applications.
Robots that are able to learn and adapt their walking will be able to operate in a much wider range of environments.

Selecting a suitable gait controller for a robot learning to walk can be very challenging, especially when targeting hardware platforms. 
A controller is often chosen early in the design process of a robot, and is used in a wide range of different evaluation budgets and environments.
Simple controllers produce gaits with a limited diversity.
More complex gait controllers are able to produce a wider range of gaits, with higher variance in performance and behaviors.

Controllers that are too complex might exhibit bootstrap problems, where the initial random population does not contain a suitable gradient towards better solutions \cite{mouret09bootstrap}.
Random solutions might also exhibit a high probability of the robot falling, making it more challenging to evolve in hardware.
Another important factor is the larger and more complex search space, which might require more evaluations to converge than practically possible without simulations \cite{tonnesfn_gecco18}.

A controller can be made simpler by embedding more prior knowledge, for instance by reducing the allowable parameter ranges of the controller. 
When the size of the search space is reduced, fewer evaluations are needed, and with more conservative parameter ranges, falling can be greatly reduced. 
Reducing the gait complexity too much, however, leaves the system with a very narrow and specialized controller that might not be able to produce gaits with the varied behaviors needed to adapt to new environments or tasks, and limitations set by human engineers might discard many near-optimal areas of the search space.

Being able to find the right complexity balance when designing a controller can be very challenging.
Any choice made early in the design process might not suit future use, and picking a single controller complexity for all different uses might end up being a costly compromise reducing performance significantly.
We have experienced this challenge in our own work where experiments are performed with a four-legged mammal-inspired robot with self-modifying morphology in both simulation and hardware \cite{tonnesfn_gecco18}.
Balancing the need for a low complexity controller when evolving morphology and control in few evaluations in hardware without falling, and evolution in complex and dynamic environments requiring exotic ways of walking in simulations, has proven impossible with our earlier controller design \cite{tonnesfn19icra}.

\begin{figure}
    \centering
    \includegraphics[width=\linewidth]{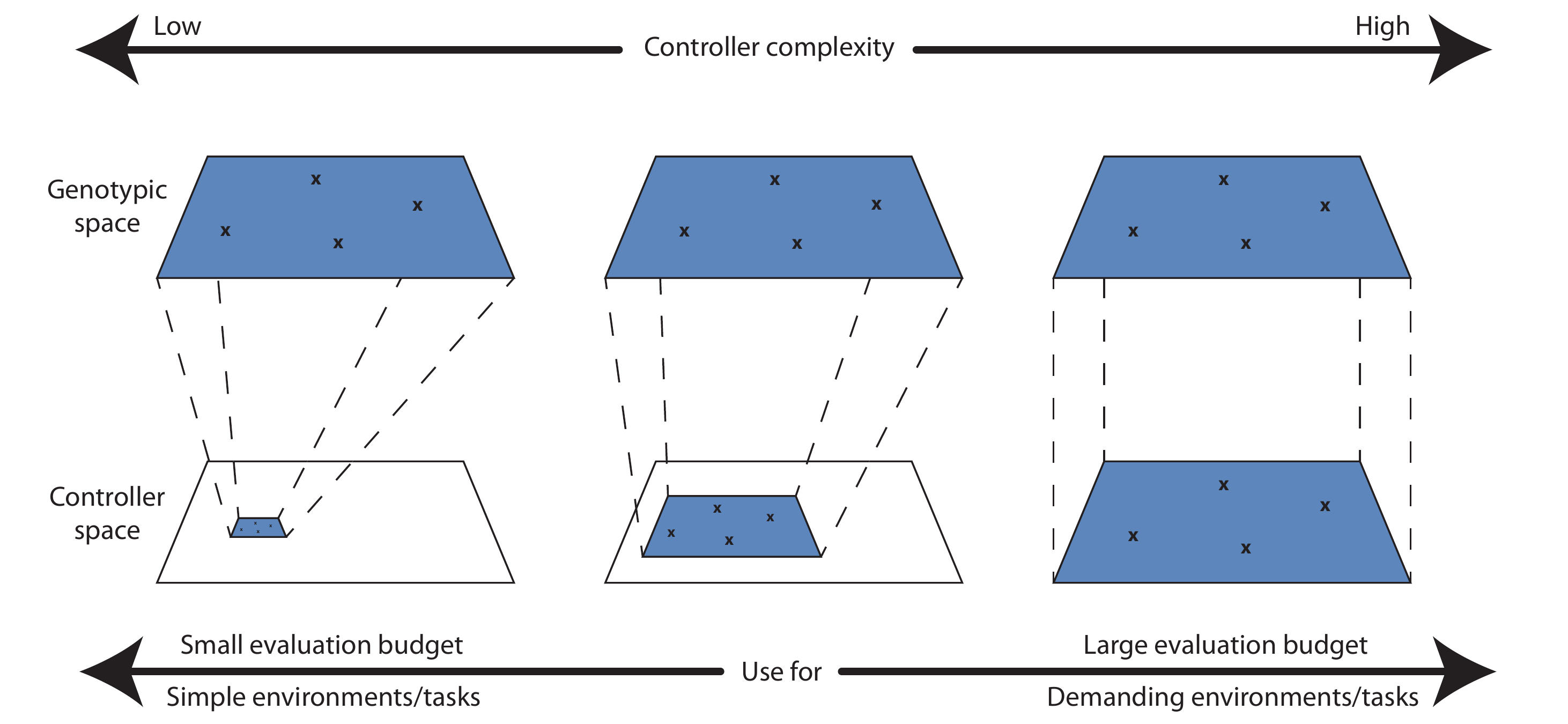}
    \caption{This diagram shows the concept of a variable complexity controller.
    The genotypic space is always the same size, but the mapping to controller space is changed by the controller complexity parameter, giving safer and more conservative gaits at lower controller complexities.}
    \label{fig.concept}
\end{figure}

In this paper, we introduce a new controller where the complexity can be set by a single parameter that addresses this limitation.
We use a dynamic genotype-phenotype mapping, illustrated in Fig. \ref{fig.concept}, where higher complexity controllers map the genotypic space to a larger controller space than lower complexity controllers.
This allows a more flexible gait either when an evaluation budget allows for longer evolutionary runs, or when the added flexibility is needed for coping with difficult environments.
Less flexible gaits can be used when there is a stricter evaluation budget, for instance in real-world experiments.
We have investigated the controller in simulation with our four-legged mammal-inspired robot, and found that  different gait complexities are optimal under different evaluation budgets. 
We also verified this through initial tests on the physical robot in the real world.
This suggests that our new controller concept will be useful for coping with the competing demands of freedom versus ease-of-learning, especially important when evolving on both virtual and real-world robots.

The contribution of this paper is as follows: We introduce the concept of a variable complexity gait controller, and show how this can be implemented for a quadruped robot. 
We then demonstrate its value through experiments in simulation, and verify the results with preliminary testing on a physical robot in the real world.

%
%

\section{Background}

Evolutionary robotics uses techniques from evolutionary computation to optimize the brain or body of a robot.
It can be used directly to improve the performance of a robot, or to study biological processes and mechanisms.
When optimizing the brain of a robot, high-level tasks like foraging, goal homing or herding can be evolved, or lower level functions like sensory perception or new walking gaits.
Optimizing the body of a robot allows adaptation to different tasks or environments, and research has shown that the complexity of evolved bodies mirror the complexity of the environments they were evolved in \cite{auerbach2014environmental}.

Several different types of optimization algorithms from evolutionary computation are used to optimize robot control. 
The most common is the Genetic Algorithm (GA) \cite{gong2010review}, which uses genetic operators like mutation and recombination to optimize gait parameters.
It is often done using multiple objectives, in many cases achieving a range of solutions with different trade-offs in conflicting objectives, including speed and stability \cite{golubovic2003ga}, or even speed, stability, and efficiency~\cite{moore2016comparison}.
Evolutionary Strategies (ES) feature self-adaptation, by adding the mutation step size to the individuals.
This has been shown to speed up the search, and in some cases outperform traditional EA approaches, when evolving quadrupedal robot gaits \cite{hebbel07es}.
Genetic Programming (GP) represents individuals as tree structures rather than vectors, and has been shown to outperform simple GA algorithms when used to evolve quadruped gaits \cite{seo10gp}.
Quality-Diversity algorithms aim to build up an archive of solutions that exhibit different behaviors or characteristics that all perform as well as possible \cite{pugh2015confronting}.
This set of diverse individuals then serves as a pool of solutions that can be searched through to find solutions to new problems, like a robot adapting to a broken leg~\cite{cully2015nature}.

Optimizing how a robot walks can be very difficult, and one of the biggest challenges is the bootstrap problem \cite{mouret09bootstrap}.
It can be very hard to start optimizing a robot gait if none of the random individuals tested initially provides a gradient towards good solutions.
This is mostly a problem when optimizing in hardware, with much harder time constraints and potential physical damage to the robot.
It can, however, also affect simulations, where initial individuals without any ability to solve a task can completely remove the selective pressure from the fitness functions needed for evolution to succeed.

There is a wide range of gait controller types used in evolutionary robotics, depending on what is being optimized. 
They are often divided into two categories, based on whether they work in the joint space, or Cartesian space \cite{GonzalezdeSantos2006}.
A gait can either be represented as a few discrete poses with trajectories generated automatically between them, or as a continuous function that specifies the position or joint angles at all times.
Some gait controllers use simple parameterized functions that control the joint space of the robot \cite{cully2015nature,tonnesfn_evostar17}.
Other gait controllers used in evolutionary experiments consist of a parameterized spline that defines each legs trajectory in Cartesian space.
Evolution optimizes either the position of the spline points directly \cite{hebbel07es}, or some higher level descriptors \cite{golubovic2003ga,tonnesfn_ices16}.
Other controllers are based on central pattern generators of different architectures and models \cite{ijspeert2008cpg}.
Some produce neural networks using techniques such as Compositional Pattern Producing Networks (CPPN), which has an inherent symmetry and coordination built-in.
This can lead to gaits far surpassing the performance of hand-designed gaits based on parameterized functions \cite{yosinski2011hyperneat}.

The field of neuro-evolution often evolves the structure of the neural networks making up the gait controller, in addition to the connection weights.
This goes against the general trend in other fields, where the complexity of gait controllers is most often kept static.
Togelius defines four different categories \cite{Togelius04neuro}.
\textit{Monolithic evolution} uses a single-layered controller with a single fitness function. \textit{Incremental evolution} in neuro-evolution has several fitness functions, but still one controller layer.
\textit{Modularised evolution} has more controller layers, but a single fitness function.
\textit{Layered evolution} uses both several controller layers, and several fitness functions.
When evolving the complexity of a network, it has been shown that new nodes should be added with zero-weights \cite{tomko2010not}, allowing evolution to gradually explore the added complexity.

%
%

\section{Implementation}

\subsection{Robot}

\begin{figure}
  \centering
  \begin{subfigure}{.48\textwidth}
    \centering
    \includegraphics[width=\linewidth]{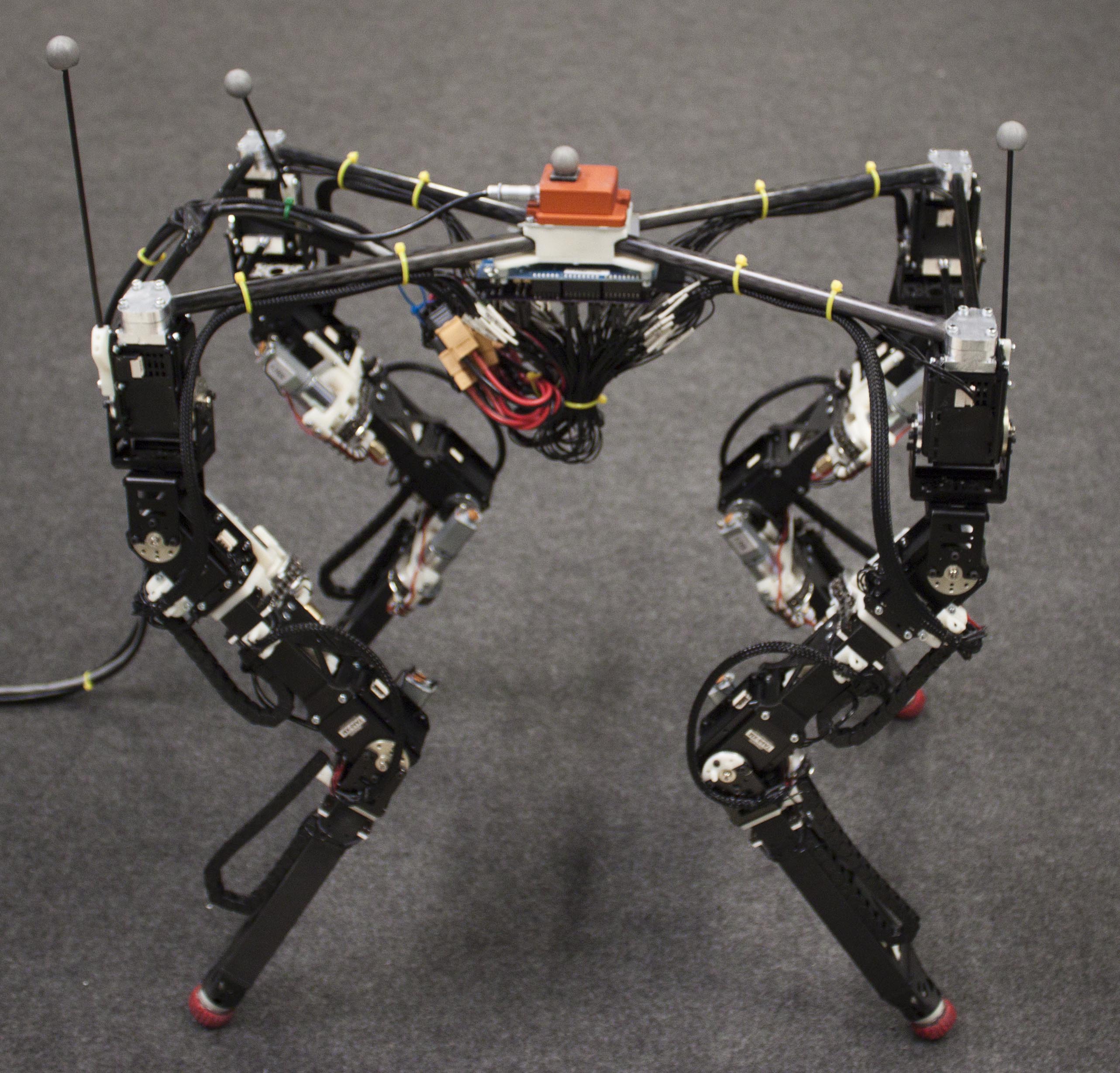}
  \end{subfigure}%
  \hspace{3mm}%
  \begin{subfigure}{.48\textwidth}
    \centering
    \includegraphics[width=\linewidth]{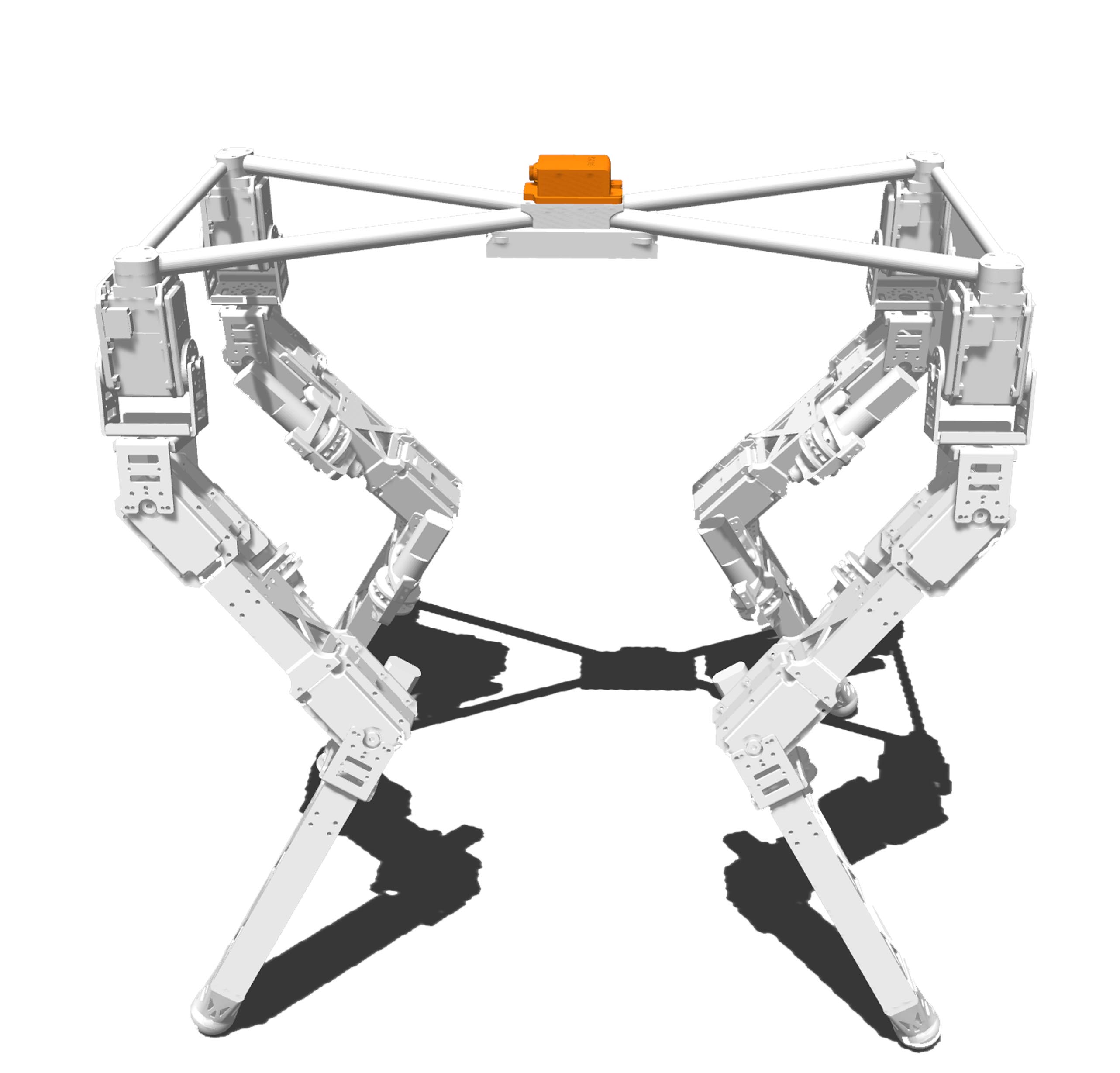}
  \end{subfigure}
  \caption{The physical robot to the left, and the simulated robot to the right.}
  \label{fig.robots}
\end{figure}

The experiments in this paper were performed on a simulated version of ``DyRET'', 
our four legged mammal-inspired robot with mechanical self-reconfiguration~\cite{tonnesfn19icra}.
The robot platform is a fully certified open source hardware project, with source and details available online\footnote{\url{https://github.com/dyret-robot/dyret_documentation}}.
We use the Robot Operating System (ROS) framework for initialization and communication, and the simulated version runs on the Gazebo physics simulator. The robot and its simulated counterpart can be seen in Fig. \ref{fig.robots}.

The robot uses Dynamixel MX-64 servos from Robotis in the hip joints, and Dynamixel MX-106 servos for the two lower joints. 
Its legs consist of two custom linear actuators each that allow reconfiguration of the leg lengths during operation.
More mechanical details can be found in our previous work~\cite{tonnesfn19icra}, and is not included here due to space constraints and the fact that we are mainly using a simulated version for our experiments.

\subsection{Control}

In our earlier experiments, we used a fairly standard parameterized spline-based gait controller working in Cartesian space.
We have used the controller for evolving both control and morphology on the physical robot, with a complex search space with many degrees of freedom.
This required us to have a low complexity controller, but that meant it was not flexible enough to give us more complex gaits when we had higher evaluation budgets, such as when using simulations.
Our goal was for the new controller to be adaptable to fit whatever needs we currently have or might have in the future, with a controller complexity that could be changed with a single parameter.

\subsubsection{The gait controller}
Since this gait is used on a physical mammal-inspired robot, the property of being learnable without excessive falling is important, and a much bigger challenge than for spider-inspired robots.
We believe that a controller operating in joint space would not allow robust enough gaits at low controller complexity for our robot, so we chose to implement it in Cartesian space.
There are many ways a gait can result in a fall, but ensuring that all legs on the ground are moving in the same direction with the same speed severely limits the chance of falling.
Complementing this with a wide leg stance gives a good base to build a parameterizable gait controller on.
Ensuring that only one leg is in the air at a time, and that the robot is always using the proper leg lift order, further helps the robot to remain stable.

\subsubsection{Leg trajectory}

\begin{figure}
    \centering
    \includegraphics[width=0.85\linewidth]{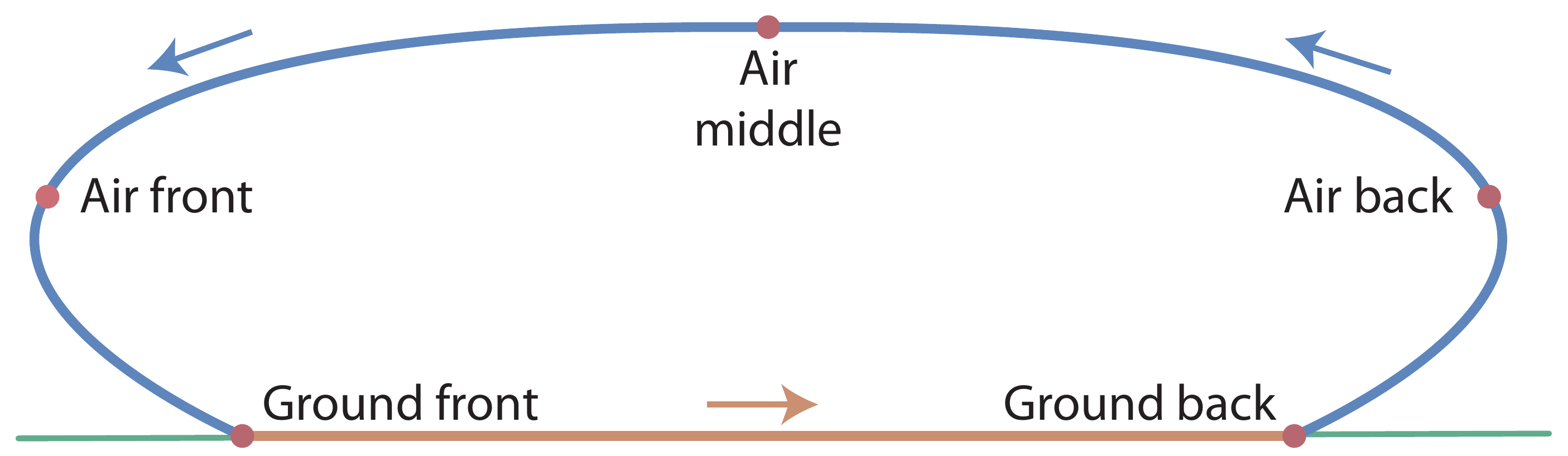}
    \caption{A simple example leg trajectory, seen from the side. The tip of the leg follows this path when the robot walks. The front of the robot is to the left.}
    \label{fig.spline}
\end{figure}

The control system uses standard inverse kinematics to get the individual joint angles from the calculated positions.
The leg trajectory is parameterized using an interpolating looping cubic Hermite spline, which intersects five control points.
A simple example trajectory can be seen in Fig. \ref{fig.spline}.
The start and end point of the spline are on the ground, while the other three points define how the leg moves forward through the air.
The leg moves in a straight line on the ground, parallel to the body of the robot, so only two parameters decide their positions.
The three points in the air are all three dimensional, with sideways movement being mirrored between left and right legs. 
This gives a total of 11 parameters that define the spline shape.

The two control points along the ground are sorted so that they always move the leg backward, while the order of the three control points in the air is chosen with an order resulting in the shortest possible spline. 
This ensures that no looping or self-intersection can happen, and allows all gait parameters to be set without constraints. 
A parameter for lift duration specifies the time the leg uses to lift back to the front, given in percentage of the gait period, while the frequency parameter gives the number of gait periods per second.

\subsubsection{Balancing wag}

In addition to positions generated for individual legs, a balancing wag is added to all legs. 
Due to the leg lift order, this can not be a simple circular motion, but needs different frequencies for the two axes.
The movement allows the robot to lean away from the leg it is currently lifting, and gives better stability.
Equation \ref{eq.wag} shows how the wag is defined, with $t$ defining the current time, and $T$ the gait period.
0.43 is a factor to offset the movement between the two wag axes to align them with the gait.
It has a phase offset ($W_\phi$) that allows for tuning to dynamic effects of the robot, while amplitude can be set separately for the two directions ($A_\text{x}$/$A_\text{y}$).
\begin{equation} \label{eq.wag}
    \begin{split}
        &W_x = \frac{A_\text{x}}{2} * tanh(3*sin( \frac{2\pi*(t+(W_\phi*T))}{T}))\\
        &W_y = \frac{A_\text{y}}{2} * tanh(3*sin( \frac{2\pi*(t+(W_\phi+0.43)*\frac{T}{2})}{\frac{T}{2}}))
    \end{split}
\end{equation}
\subsubsection{Complexity scaling}
The complexity of the controller can be modified by a single parameter, from 0 to 100\%. 
There are many ways to provide a scaling of the complexity of the controller, but we chose to implement this using a dynamic genotype-phenotype mapping that varies the range of gait parameters linearly with the controller complexity.
All controller parameters have a center value, that together with the minimum range gives the allowable range at controller complexity 0\%. 
These have been chosen so they represent a very conservative and safe controller that should work well in most conditions, based on traditional robotics techniques and earlier experience with the robot. 
Using a more complex controller by allowing a large range of values, however, allows the controller to deviate from the safe values and into the more extreme values often needed for more complex environments or tasks.
Parameters controlling the spline shape can be seen in Table \ref{table.splineParams}, with high-level gait parameters in Table \ref{table.gaitParams}.

\begin{table}
    \centering
    \caption{Parameters and ranges defining the spline shape}
    \begin{tabular}{ l | r r r r }
        \hline
        Control point & Minimum & Maximum &  \hspace{1mm}Default value &  \hspace{1mm}Minimum range \\
        \hline
        Ground front & -150 & 150 &   50 & 50 \\
        Ground back  & -150 & 150 & -100 & 50 \\
        Air 1 & \hspace{2mm}[-25, -150, 10] & \hspace{2mm}[25, 150, 80] & [0, 75, 30] & [0, 50, 10] \\
        Air 2 & [-25, -150, 10] & [25, 150, 80] & [0, 0, 50]  & [0, 0, 10] \\
        Air 3 & [-25, -150, 10] & [25, 150, 80] & [0, -75, 50]  & [0, 50, 10] \\
        \hline
    \end{tabular}
    \label{table.splineParams}
\end{table}
\begin{table}
    \centering
    \caption{Parameters and ranges of gait parameters.}
    \begin{tabular}{ l | r r r r }
        \hline
        Parameter & Minimum & Maximum & Default value & Minimum range \\
        \hline
        Wag phase       & -$\pi$/2 & $\pi$/2 &     0 &  0.2 \\
        Wag amplitudes  &        0 &      50 &     0 &    5 \\
        Lift duration   &     0.05 &    0.20 & 0.175 & 0.05 \\
        Frequency       &     0.25 &    1.5  &     - &    - \\
        \hline
    \end{tabular}
    \label{table.gaitParams}
\end{table}

\begin{figure}
    \centering
    \makebox[20pt]{\raisebox{40pt}{\rotatebox[origin=l]{90}{\hspace{-125px}50\%\hspace{32px}25\%\hspace{32px}0\%}}}%
    \begin{subfigure}[t]{0.47\textwidth}
        \includegraphics[width=\textwidth]{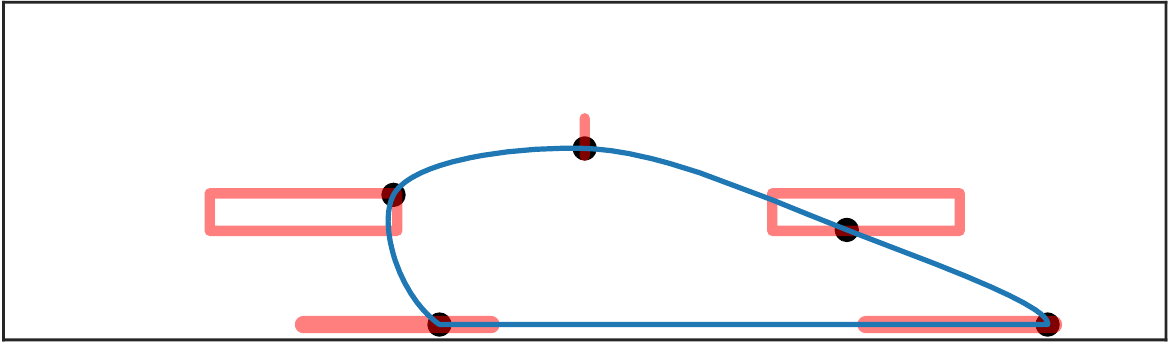}
        \includegraphics[width=\textwidth]{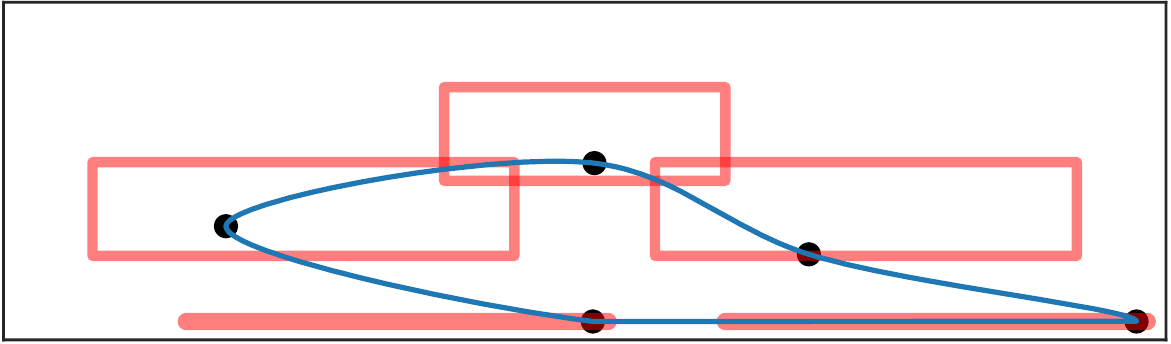}
        \includegraphics[width=\textwidth]{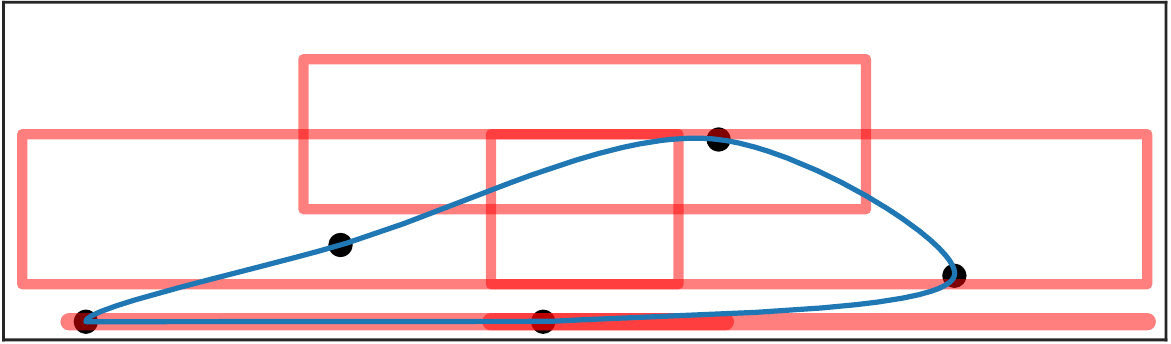}
    \end{subfigure}\hfill
    \begin{subfigure}[t]{0.47\textwidth}
        \includegraphics[width=\textwidth]{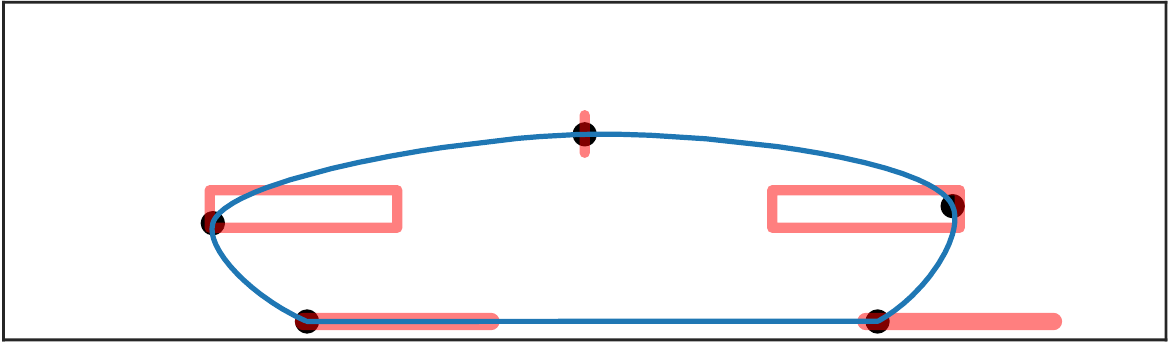}
        \includegraphics[width=\textwidth]{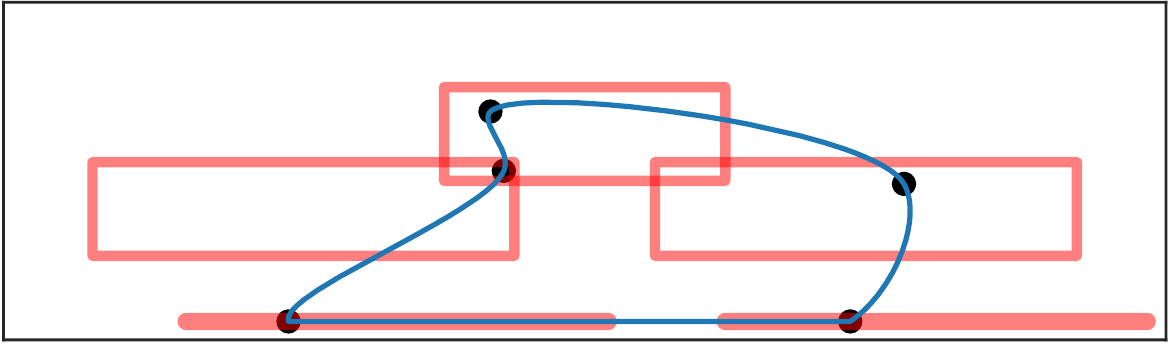}
        \includegraphics[width=\textwidth]{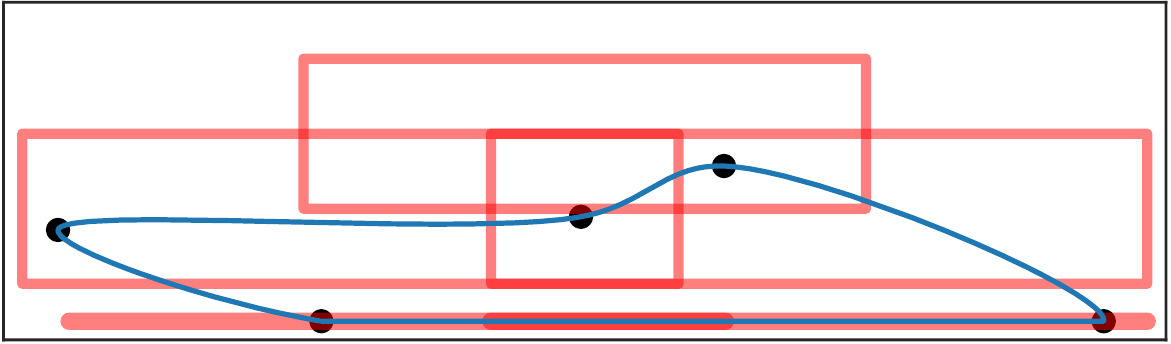}
    \end{subfigure}\hfill
    \caption{Examples of leg trajectory splines generated at different gait complexities. These are seen from the side of the robot, with the front of the robot to the left of the plot. The red boxes show the range of possible control point positions.}
    \label{fig.splineExamples}
\end{figure}

Examples of splines with different gait complexities can be seen in Fig.~\ref{fig.splineExamples}.
For complexities of 0, the splines are fairly conservative, but even though the parameter ranges are low, they do show some variation in their basic shapes.
The higher complexity gaits have spline shapes that are much more unconventional, though sorting the control points to minimize spline length does remove self-intersections to keep all trajectories feasible.
Please note that the plot shows the commanded position to the robot, and that the actual leg trajectory can be very different than commanded, due to the mechanical and control properties of the actuators, and the dynamics of the system. 
Very complex shapes that appear unintuitive for human engineers might end up giving much smoother and higher performing gaits in the real world than expected.

\subsection{Evolutionary setup}
Here we describe the setup we used for evolving the controllers, as well as how we evaluated them. We evolved controllers for both stable and fast forward walking on flat ground.

\subsubsection{Evolutionary algorithm and operators}
We used the NSGA-II evolutionary algorithm, running on the Sferes2 evolutionary framework.
We chose this algorithm since we are optimizing both speed and stability, but would not like to choose the specific trade-off between the two objectives before optimization. 
NSGA-II features a mechanism to increase the crowding distance in the Pareto front, which gives a wide range of trade-offs to pick from.

Gaussian mutation was used with a mutation probability of 100\% and a sigma of 1/6.
No recombination operators were used.

Early experimentation showed a big difference in the number of evaluations before convergence for different controller complexities, which suggested the need for different population sizes.
We tested a range of different population sizes at the minimum and maximum complexity, as well as a few points in between, and found that a population of eight at zero complexity, and 64 at full complexity worked best.
Population sizes for all intermediary complexities were set linearly, and rounded to the nearest power of two.
Tests showed that runs at all gait complexities converge to a satisfactory degree after 8192 evaluations.

We performed 25 runs for each controller complexity in simulations to gain a good estimate of the performance.
Each simulated run took about 11 hours, and we used about 10,000 CPU core hours on the simulation for the experiments featured in the paper.
Experiments in the real world take a lot longer, so we  only performed three runs for each controller complexity, as the experiment only serves as a preliminary test to see confirm simulated results in  the real world.

\subsubsection{Fitness objectives}
We used both speed and stability as our fitness measurements.
Speed was calculated as the distance between start and stop position, divided by the evaluation time, as seen in equation \ref{eq.fitSpeed}. 
Distance was measured using motion capture equipment in the real world, and extracted directly in simulation.
Only the speed straight forward was used, so we filtered out any sideways movement by only measuring position in the forward axis.
Stability was calculated with a weighted sum of the variance in acceleration and orientation. 
The full fitness function for stability can be seen in equation \ref{eq.fitStab}, where \textit{acc} are samples from the accelerometer, \textit{ang} are samples from the orientation output of the Attitude and Heading Reference System (AHRS), \textit{i} is the sample index, and \textit{j} is the axis of the sample.
The Xsens Mti-30 AHRS was used on the physical robot, and a virtual version of the same was used in simulation.

\begin{equation}
  F_{speed} = \frac{\lVert P_{end} - P_{start} \rVert}{time_{end} - time_{start}}
  \label{eq.fitSpeed}
\end{equation}

\begin{equation*}
  G(A_{j}) = \sqrt{\frac{1}{n}\sum\limits_{i=1}^{n} (A_{j,i}^2-\overline{A_{j}}^2)}\\
\end{equation*}

\begin{equation}
  F_{stability} = -\left(\alpha * \sum\limits^{axes}{G(Acc_{axis})} + \sum\limits^{axes}{G(Ang_{axis})} \right)
  \label{eq.fitStab}
\end{equation}

\subsubsection{Evaluation}

We ran all our simulations on the Gazebo physics simulator.
Each gait was evaluated in simulation by walking forwards 1 meter, with a timeout of 10 seconds.
The position and pose of the robot were reset between all evaluations.

Evaluating and comparing the performance of different optimization runs can be challenging when doing multi-objective optimization.
This is especially true when using an algorithm like NSGA-II, that has a mechanism for stretching out the Pareto front, making it hard to compare the two objectives separately.
Therefore, we instead looked at the hypervolume~\cite{hypervolume} when comparing populations.
The hypervolume measures the volume (or area, in the case of two objectives) of the dominated part of the objective space.
The lower bound of stability was set to -1 for the hypervolume calculation, while speed was capped to 0 m/min.

%
%

\section{Experiments and results}

We present the results of experiments in simulation and on a real-world robot.
These experiments are simplified and performed with as many variables removed as possible. The robot's task is to walk straight forward, and the environment is a flat surface with medium friction, both in simulation and the real world.

\subsection{Finding the maximum needed complexity}

First, we wanted to investigate whether there is a maximum controller complexity needed for the environment and task we are using.
Since neither is very challenging, we do not expect the need for very complex controllers.
We ran full evolutionary runs at a range of gait complexities.

\begin{figure}
    \centering
    \includegraphics[width=\linewidth]{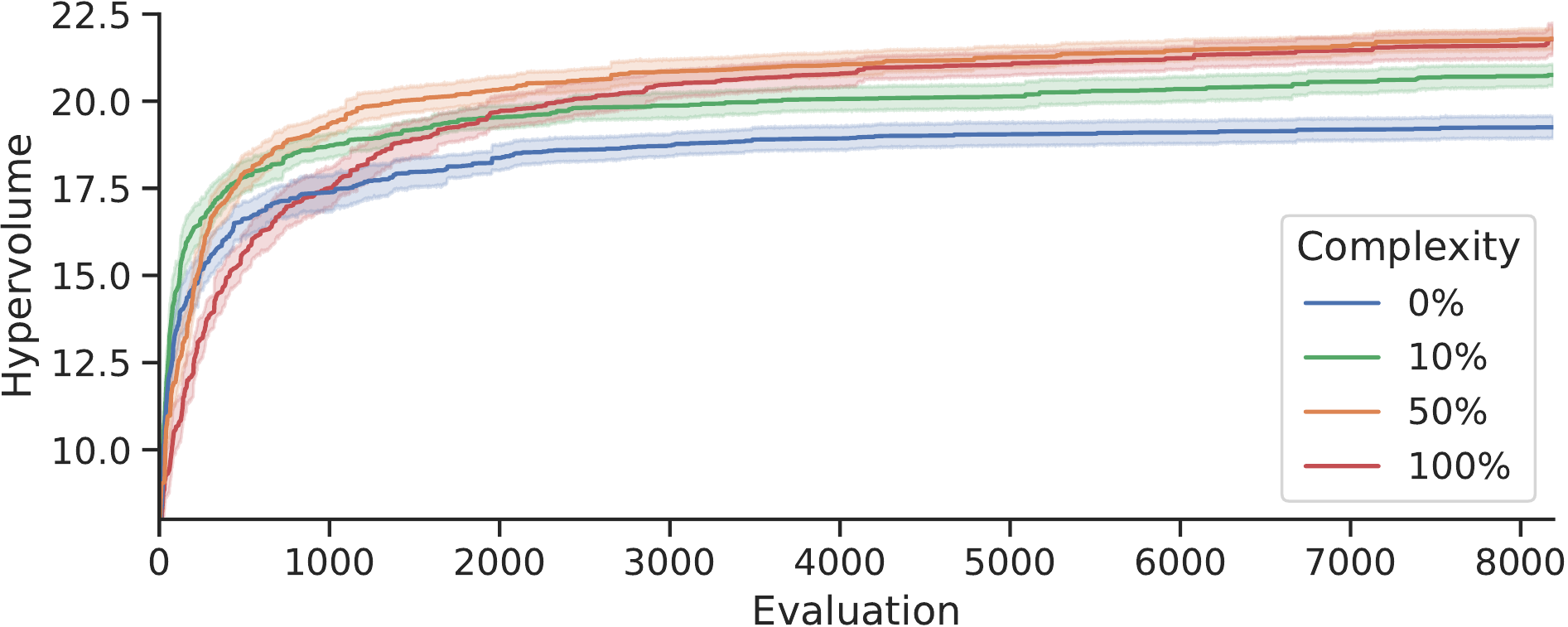}
    \caption{Hypervolume from evolutionary runs with selected gait complexities. The solid lines show the means, with 95\% confidence interval in the shaded areas.}
    \label{fig.hypervolumes_8192}
\end{figure}

Fig. \ref{fig.hypervolumes_8192} shows how the hypervolume progresses over evaluations.
This shows that the lower complexity controllers converge quicker, but are not able to achieve the same performance as the higher complexity controllers.
The 50\% and 100\% complexity controllers end up with the same performance, though the 100\% complexity controller takes considerably longer to converge.

\begin{figure}
  \centering
  \hspace{-3mm}
  \includegraphics[width=\textwidth]{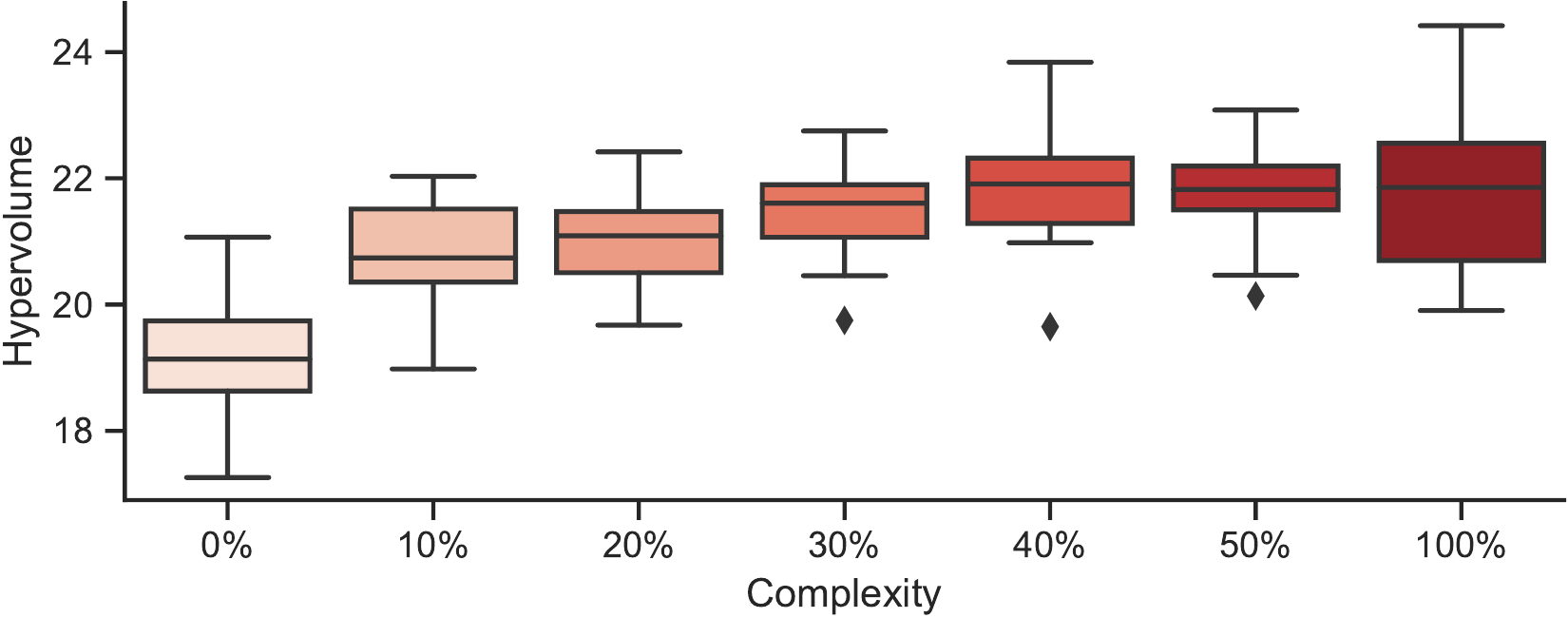}
  \caption{Hypervolume for runs with gait controller complexities ranging from 0\% to 100\%, showing details from the end results of the optimization process.}
  \label{fig.hypervolumeboxplot}
\end{figure}

The details of the last evaluations of the runs are better illustrated with the boxplots, seen in Fig.~\ref{fig.hypervolumeboxplot}.
These show the distribution of the hypervolumes achieved at the end of all the optimization runs.
The hypervolume improves for gait complexities from 0\% to 40\%, but there is no improvement between 40\% and 50\%.
100\% complexity has a wider spread than the others, which might be beneficial in some applications, but the median performance is no better than the 40-50\% complexity.

\subsection{Complexity for different evaluation budgets}

A potentially rewarding feature of controlling the complexity of the gait is the ability to adapt it to a specific evaluation budget. 
The price of computational resources is decreasing, enabling a large number of evaluations in simulation. 
Hardware experiments, however, are limited by the number of robots that can be built, maintained, and supervised during experiments.
Evaluation is therefore much more expensive for hardware experiments than for simulations, and this gap will only increase.

For this investigation, we have selected a range of different evaluation budgets to test. 
We have previously used 64 and 128 evaluations in our hardware experiments \cite{tonnesfn_ices16,tonnesfn_gecco18}, and 512, 2048 and 8192 evaluations gives a range more appropriate for simulation experiments.

\begin{figure}
  \centering
  \includegraphics[width=\textwidth]{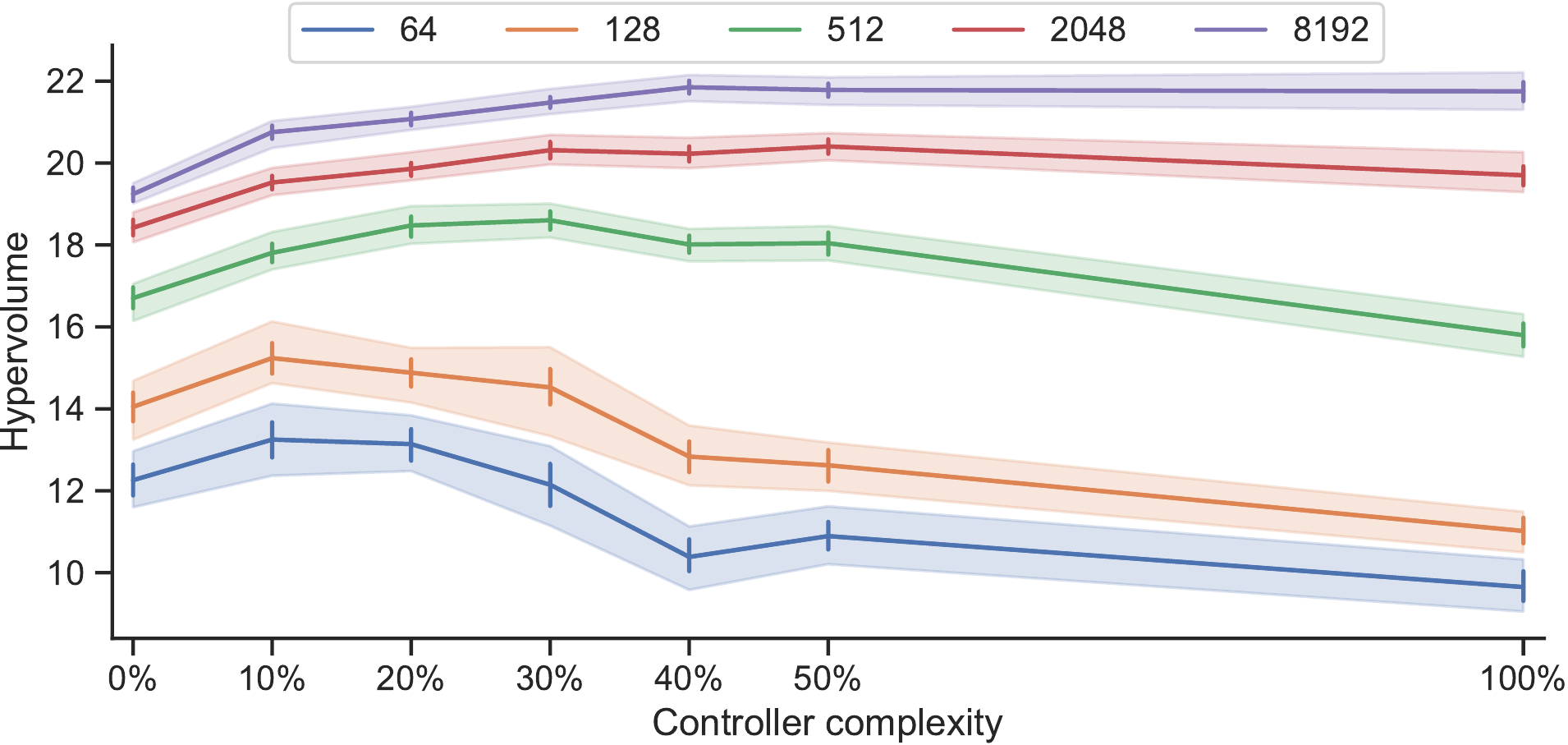}
  \caption{This figure shows how different controller complexities affects achievable hypervolume for different evaluation budgets. The vertical lines show the standard deviation, while the shaded areas show the 95\% confidence intervals.}
  \label{fig.complexityhypervolume}
\end{figure}

Fig. \ref{fig.complexityhypervolume} shows how the controller complexity affects achieved hypervolume for the different budgets.
For the shortest two simulation cases, with 64 and 128 evaluations, hypervolume is highest at 10\% complexity.
Budgets 512 and 2048 achieve the best performance around 30\%, while the long simulation case performs best at 40\%-100\%.

\subsection{Analyzing resulting populations}

\begin{figure}
    \centering
    \begin{subfigure}[b]{0.95\textwidth}
        \includegraphics[width=\textwidth]{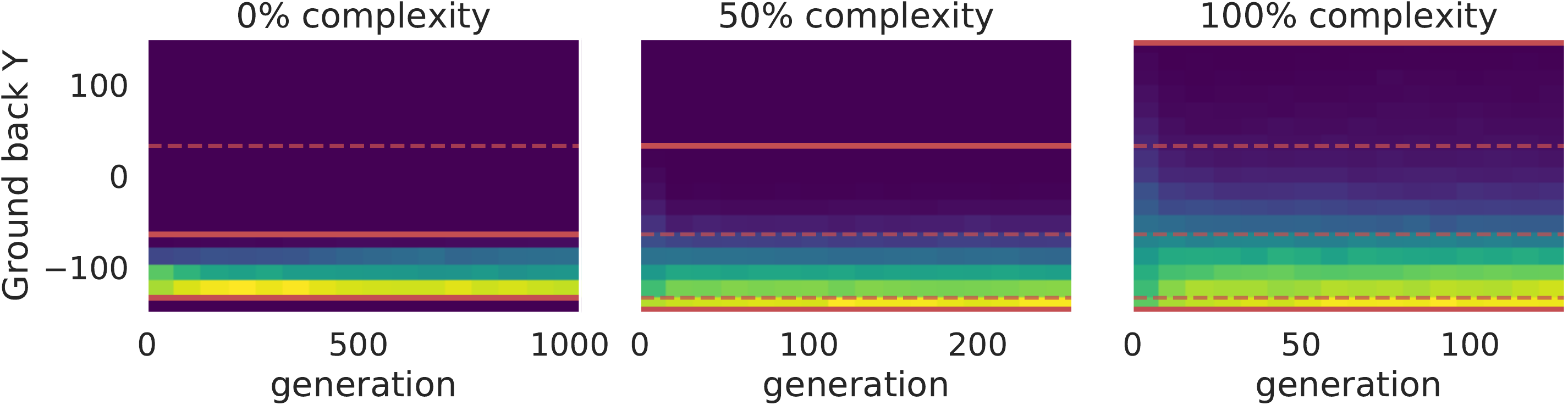}
        \caption{The $y$ position of the second ground control point in the spline.}
        \label{fig.hist2d_p1_y}
        \vspace{1.5mm}
    \end{subfigure}
    \begin{subfigure}[b]{0.95\textwidth}
        \includegraphics[width=\textwidth]{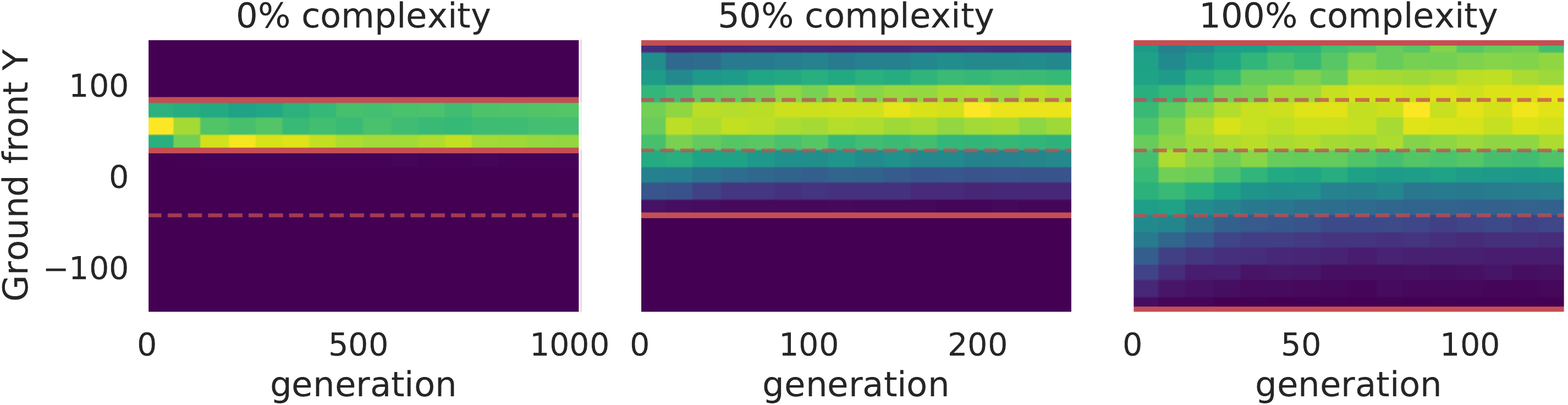}
        \caption{The $y$ position of the first ground control point in the spline.}
        \label{fig.hist2d_p0_y}
        \vspace{1.5mm}
    \end{subfigure}
    \begin{subfigure}[b]{0.95\textwidth}
        \includegraphics[width=\textwidth]{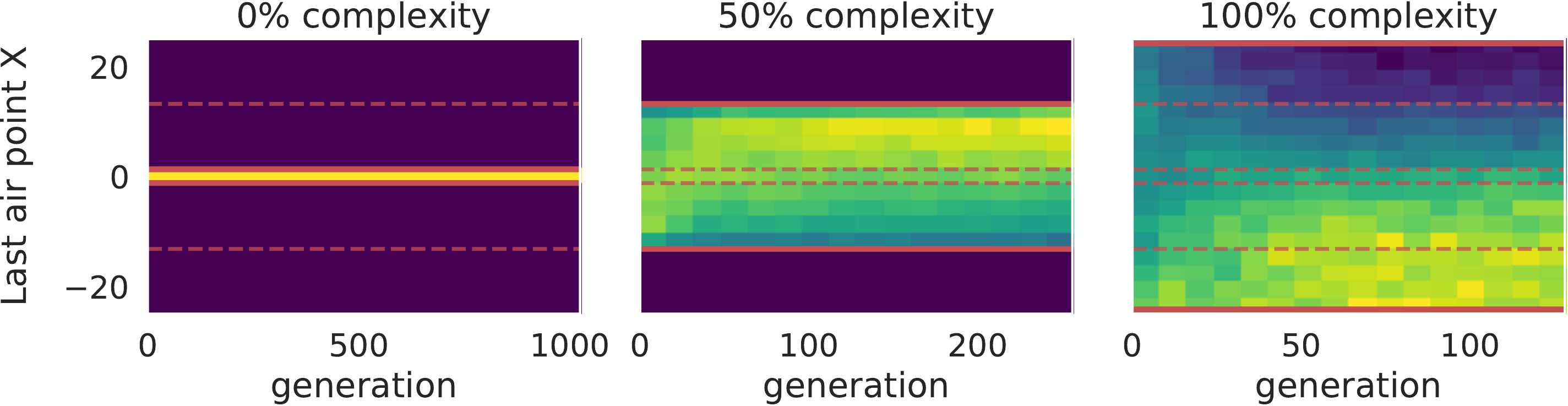}
        \caption{The $x$ position of one of the air control points in the spline.}
        \label{fig.hist2d_p4_x}
        \vspace{1.5mm}
    \end{subfigure}
    \begin{subfigure}[b]{0.95\textwidth}
        \includegraphics[width=\textwidth]{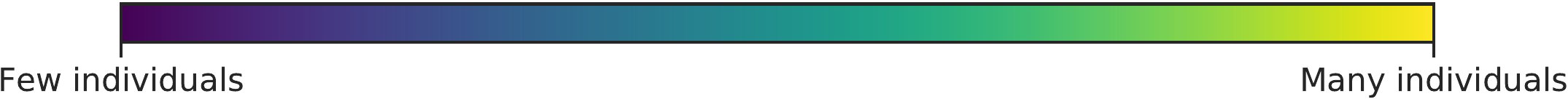}
    \end{subfigure}
    \caption{Values of a select few parameters throughout the optimization run. The solid red lines show the range of the parameters, and the dashed red lines trace the range from the other complexities to ease comparison.}
    \label{fig.hist2d}
\end{figure}

Figure \ref{fig.hist2d} shows which parameters are tested at various parts of the search. 
Some parameters, like the $y$ position of the back ground control point, end up close to their conservative estimate, and do not exploit their additional freedom from the higher complexity in our simple experiments, as seen in Figure \ref{fig.hist2d_p1_y}.
Other parameters, like the $y$ position of the front ground control point, do use more of their available range, although it is still close to its original estimate.
In Figure \ref{fig.hist2d_p4_x}, the search with 50\% controller complexity seems to maximize the $x$ position of the third air control point in the spline, while with the whole area available in the 100\% complexity controller, it ends up minimizing it.

\subsection{Initial hardware testing}

\begin{figure}
  \centering
  \includegraphics[width=\textwidth]{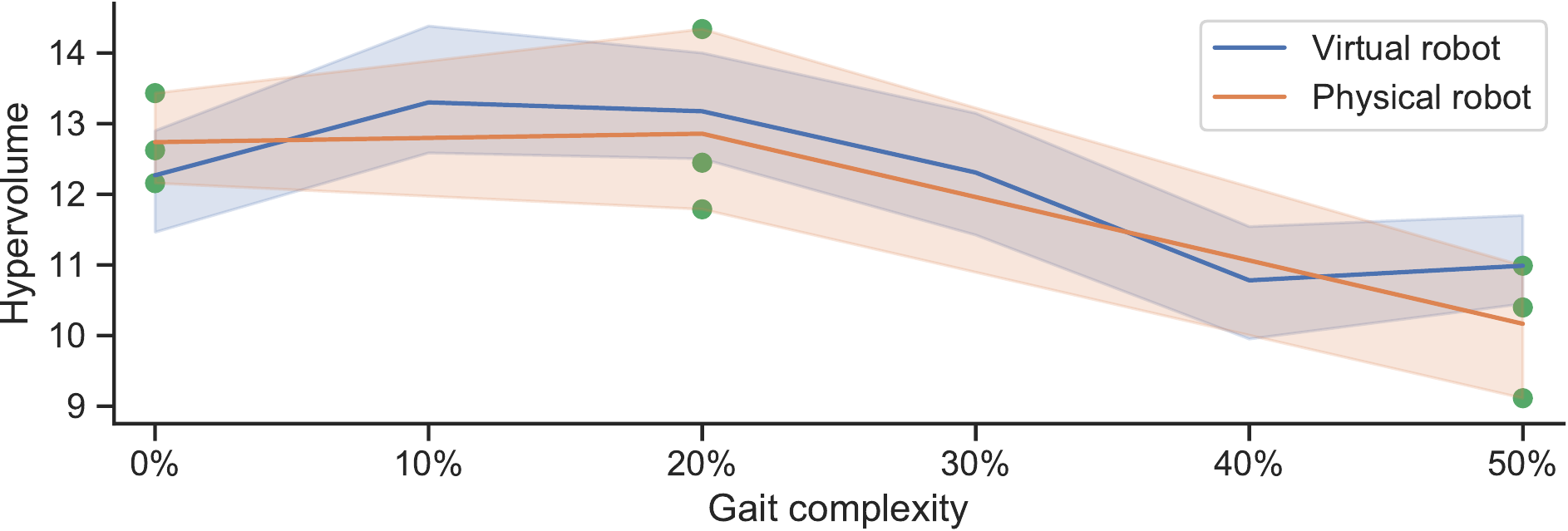}
  \caption{This figure shows the performance on the real robot, compared to the simulated version seen in Fig. \ref{fig.complexityhypervolume}. The added green dots are the resulting hypervolume from each of the runs in hardware.}
  \label{fig.complexityhypervolume_hw}
  \vspace{-5mm}
\end{figure}

We also did evolutionary runs using this new controller on the physical robot in the real world with 64 evaluations per run, using eight generations of eight individuals.
We decided to test a controller complexity of 0\%, as well as 50\%, which is the highest complexity we were confident in using on the physical robot without excessive risk of physical damage to the system.
We also tested 20\%, which gives us another data point between these, and was among the two top performing complexities in simulation with this evaluation budget.
The results can be seen in Fig. \ref{fig.complexityhypervolume_hw}, where we can see the same general trends as in the simulator.
Controller complexities 0\% and 20\% both did well, and we are not able to separate the two with the limited number of evaluations we were able to do in hardware.
50\% controller complexity, however, does considerably worse than the other two, just as we saw in simulation.

\section{Discussion}

The performance differences in Fig. \ref{fig.complexityhypervolume} suggest that choosing the right controller complexity for an evaluation budget can be very important, especially when that budget is small.
Lower complexity controllers fall less, so if optimization is done in hardware, this could also be taken into account when deciding on the complexity.
We did a simple grid-search for our experiments since we were only investigating the controller, but more advanced search algorithms could be be performed to further optimise the choice of complexity.

We used different population sizes when evolving with different complexities in our experiments. 
Our controller was designed to be evolved with evaluation budgets as small as 32 or 64 evaluations when doing real world experiments, and with budgets larger than 8192 when evolving in simulation.
Limiting the population size to the smallest budget would give a very unrealistic measurement of performance for the larger budgets, and thus we chose suitable population sizes for the different complexities through simple trial and error.
This does obfuscate the results to a degree, but we feel this gives the most fair comparison.
The evolutionary operators would likely also be slightly different, but they were kept the same as they affect the search to a much smaller degree.

The parameter for the $x$ position of the third air control point, seen in Fig. \ref{fig.hist2d_p4_x}, seems to be maximized at 50\% complexity, but be minimized at 100\% complexity. 
This is most likely due to interactions between different parameters.
At half complexity, the optimal value might be towards the top of the parameter range.
At full complexity, however, new ranges for the other parameters are opened up, allowing better performance for lower parts of the range.

The choice of centers and minimum ranges for each gait parameter greatly affect the performance of lower complexity gait controllers.
The choice should be based on conservative values that are assumed to work sufficiently in all environments, not on optimal values for a single environment.
Evolution is often used to adapt to changes in environments or tasks.
If the centers and ranges were chosen after optimal solutions were found, they would most likely not perform well when things change, and one might as well just select the top performing individuals from simulation directly.
In our case, we selected these values before doing the optimization, and several are far from optimal.
This can be seen in Fig. \ref{fig.hypervolumeboxplot}, where the performance of the low complexity controller is much worse than for the higher complexity ones.
This is by design, as safe and conservative parameters that work for all environments rarely do very well in any of them.

The choice of maximum ranges also affect the outcome, but not to as high a degree as the center and minimum range.
Limiting the ranges too much means the controller will never be able to achieve the potential increase in performance from that specific controller feature.
Having ranges that are too large, with values that will never be optimal under any circumstance,  serves to slow down the search, and waste time and resources.
A good optimization algorithm that is not getting stuck in early local optima, however, should be able to converge outside these infeasible areas.
We therefore recommend anyone implementing this type of controller to spend some time choosing parameter centers and minimum ranges to be conservative and safe, but not be afraid to overshoot a bit on the maximum allowable range, as the consequence of choosing ranges that are too narrow is far worse than selecting too high.

Figure \ref{fig.complexityhypervolume_hw} shows the results from the testing in hardware.
We are unable to say anything definitive with the results due to the low number of evaluations and the relatively high degree of noise, but it does support what we found in simulation.
Not only did the 50\% controller complexity perform worse, like predicted in simulation, but we also experienced qualitatively more extreme gaits, and actually had to pause the evolutionary runs at several times to repair the robot after damage.
We also experienced several falls with the 50\% controller complexity, but no falls or damage at the two lower complexities, supporting our original assumption that the gait values were conservative and safe.

We consider this type of controller to be very useful for researchers doing gait optimization in the real world on physical robots, as the reality gap can often times make it impractical or impossible to directly use individuals from simulation in the real world.
Simulations can be used to find approximate upper bounds of the needed complexity as we saw in Fig. \ref{fig.hypervolumeboxplot}, but even more useful is being able to tune the complexity to the limited evaluation budget used in hardware, as seen in Fig. \ref{fig.complexityhypervolume}. 
We also expect that more demanding or dynamic environments and tasks might be able to exploit higher complexities better than what we experienced in our experiments, which only included forward walking in straight lines on even terrain.

%
%
\section{Conclusion and future work}

In this paper, we introduced our new gait controller with variable complexity.
We tested the controller in simulation, and found that different gait complexities are optimal for different evaluation budgets.
We also did preliminary tests on a physical robot in the real world that supported our findings. 
Being able to change the controller complexity allows a researcher to use less complex controllers when optimizing gait on a physical robot, and increase the complexity when needed for demanding environments, or when doing longer optimization in simulations.


One natural extension of our work is to use our variable complexity controller in incremental evolution. 
Since this controller offers a continuous complexity parameter, the difficulty can be gradually increased for each generation.
Since an increase in difficulty follows a known set of rules, all individuals can keep their phenotypic values between generations, even when parameter ranges are expanded.
This allows evolution to gradually explore the added complexity, in the same way that has been shown to be optimal for neuro-evolution \cite{tomko2010not}.
The controller complexity can also be changed during the evolutionary process as part of evolutionary strategies, or be controlled during robot operation as part of lifelong learning.

We have only tested this controller in a single environment in simulation where complexities over 50\% were not needed.
It would be interesting to test it in more challenging and dynamic environments to see if controllers with higher complexities are able to use the increased parameter ranges to actually increase performance.
Doing a more thorough investigation into the parameters selected might yield ranges or values that act limiting on the fully complex controller, and would allow even more flexible gaits.
Analyzing the individual leg trajectories evolved would also be interesting, and could shed light on the matter from a different perspective.
Investigating how evolutionary meta-parameters interact with the complexity would be interesting, including population size and evolutionary operators.
Adding sensing and allowing the robot to choose which complexity is needed for its current environment is also worth exploring.

%
%

\bibliographystyle{splncs03}
\bibliography{nygaard}

\begin{thebibliography}{10}
\providecommand{\url}[1]{\texttt{#1}}
\providecommand{\urlprefix}{URL }

\bibitem{mouret09bootstrap}
Mouret, J., Doncieux, S.: Overcoming the bootstrap problem in evolutionary
  robotics using behavioral diversity. In: 2009 IEEE Congress on Evolutionary
  Computation. pp. 1161--1168 (May 2009)

\bibitem{tonnesfn_gecco18}
Nygaard, T.F., Martin, C.P., Samuelsen, E., Torresen, J., Glette, K.:
  Real-world evolution adapts robot morphology and control to hardware
  limitations. In: Proceedings of the Genetic and Evolutionary Computation
  Conference. ACM (2018)

\bibitem{tonnesfn19icra}
Nygaard, T.F., Martin, C.P., Torresen, J., Glette, K.: Self-modifying
  morphology experiments with dyret: Dynamic robot for embodied testing. In:
  2019 IEEE International Conference on Robotics and Automation (ICRA) (May
  2019)

\bibitem{auerbach2014environmental}
Auerbach, J.E., Bongard, J.C.: Environmental influence on the evolution of
  morphological complexity in machines. PLoS computational biology  10(1)
  (2014)

\bibitem{gong2010review}
Gong, D., Yan, J., Zuo, G.: A review of gait optimization based on evolutionary
  computation. Applied Computational Intelligence and Soft Computing  (2010)

\bibitem{golubovic2003ga}
Golubovic, D., Hu, H.: Ga-based gait generation of sony quadruped robots. In:
  Proc. 3th IASTED Int. Conf. Artificial Intelligence and Applications (AIA)
  (2003)

\bibitem{moore2016comparison}
Moore, J.M., McKinley, P.K.: A comparison of multiobjective algorithms in
  evolving quadrupedal gaits. In: International Conference on Simulation of
  Adaptive Behavior. pp. 157--169. Springer (2016)

\bibitem{hebbel07es}
Hebbel, M., Nistico, W., Fisseler, D.: Learning in a high dimensional space:
  Fast omnidirectional quadrupedal locomotion. In: Lakemeyer, G., Sklar, E.,
  Sorrenti, D.G., Takahashi, T. (eds.) RoboCup 2006: Robot Soccer World Cup X
  (2007)

\bibitem{seo10gp}
Seo, K., Hyun, S., Goodman, E.D.: Genetic programming-based automatic gait
  generation in joint space for a quadruped robot. Advanced Robotics  24(15)
  (2010)

\bibitem{pugh2015confronting}
Pugh, J.K., Soros, L.B., Szerlip, P.A., Stanley, K.O.: Confronting the
  challenge of quality diversity. In: Proceedings of the 2015 Annual Conference
  on Genetic and Evolutionary Computation. pp. 967--974. ACM (2015)

\bibitem{cully2015nature}
Cully, A., Clune, J., Tarapore, D., Mouret, J.B.: Robots that can adapt like
  animals. Nature  521(7553),  503 (2015)

\bibitem{GonzalezdeSantos2006}
de~Santos, P.G., Garcia, E., Estremera, J.: Quadrupedal locomotion: an
  introduction to the control of four-legged robots. Springer Science \&
  Business Media (2007)

\bibitem{tonnesfn_evostar17}
Nygaard, T.F., Samuelsen, E., Glette, K.: Overcoming initial convergence in
  multi-objective evolution of robot control and morphology using a two-phase
  approach. In: Squillero, G., Sim, K. (eds.) Applications of Evolutionary
  Computation. pp. 825--836. Springer International Publishing (2017)

\bibitem{tonnesfn_ices16}
Nygaard, T.F., Torresen, J., Glette, K.: Multi-objective evolution of fast and
  stable gaits on a physical quadruped robotic platform. In: 2016 IEEE
  Symposium Series on Computational Intelligence (SSCI) (Dec 2016)

\bibitem{ijspeert2008cpg}
Ijspeert, A.J.: Central pattern generators for locomotion control in animals
  and robots: a review. Neural networks  21(4),  642--653 (2008)

\bibitem{yosinski2011hyperneat}
Yosinski, J., Clune, J., Hidalgo, D., Nguyen, S., Zagal, J.C., Lipson, H.:
  Evolving robot gaits in hardware: the hyperneat generative encoding vs.
  parameter optimization. In: ECAL. pp. 890--897 (2011)

\bibitem{Togelius04neuro}
Togelius, J.: Evolution of a subsumption architecture neurocontroller. J.
  Intell. Fuzzy Syst.  15(1),  15--20 (Jan 2004)

\bibitem{tomko2010not}
Tomko, N., Harvey, I.: Do not disturb: Recommendations for incremental
  evolution. In: Proceedings of ALIFE XII, the 12th International Conference on
  the Synthesis and Simulation of Living Systems (2010)

\bibitem{hypervolume}
Knowles, J.D., Corne, D.W., Fleischer, M.: Bounded archiving using the lebesgue
  measure. In: The 2003 Congress on Evolutionary Computation. vol.~4 (Dec 2003)

\end{thebibliography}

\end{document}